\documentclass[conference]{IEEEtran}
\ifCLASSINFOpdf
\else
\fi
\usepackage{graphicx}
\usepackage{amsmath, amssymb}
\usepackage{siunitx}
\usepackage{svg}
\usepackage[utf8]{inputenc}
\usepackage[vietnam]{babel}

\begin{document}
\title{From Universal Language Model to Downstream Task: Improving RoBERTa-Based Vietnamese Hate Speech Detection}
\author{
    \IEEEauthorblockN{Quang Huu Pham\IEEEauthorrefmark{1}, Viet Anh Nguyen\IEEEauthorrefmark{1}, Linh Bao Doan, Ngoc N. Tran}
    \IEEEauthorblockA{R\&D Lab, Sun Asterisk Inc \\
    \{pham.huu.quang, nguyen.viet.anhd, doan.bao.linh, tran.ngo.quang.ngoc\}@sun-asterisk.com}
    
    \\
    \IEEEauthorblockN{Ta Minh Thanh\IEEEauthorrefmark{2}}
    \IEEEauthorblockA{Le Quy Don Technical University, 236 Hoang Quoc Viet, Bac Tu Liem, Ha Noi \\
    thanhtm@mta.edu.vn}
    
    \IEEEauthorrefmark{1} Equal contribution
    \IEEEauthorrefmark{2} Corresponding author
}
\maketitle

\begin{abstract}
Natural language processing (NLP) is a fast-growing field of artificial intelligence. Since the Transformer \cite{vaswani2017attention} was introduced by Google in 2017, a large number of language models such as BERT, GPT, and ELMo have been inspired by this architecture. These models were trained on huge datasets and achieved state-of-the-art results on natural language understanding. However, fine-tuning a pre-trained language model on much smaller datasets for downstream tasks requires a carefully-designed pipeline to mitigate problems of the datasets such as lack of training data and imbalanced data. In this paper, we propose a pipeline to adapt the general-purpose RoBERTa language model to a specific text classification task: Vietnamese Hate Speech Detection. We first tune the PhoBERT\footnote{PhoBERT is a pre-trained RoBERTa model which is known as a state-of-the-art language model for Vietnamese provided by VinAI research: https://github.com/VinAIResearch/PhoBERT}\cite{nguyen2020phobert} on our dataset by re-training the model on the Masked Language Model (MLM) task; then, we employ its encoder for text classification. In order to preserve pre-trained weights while learning new feature representations, we further utilize different training techniques: layer freezing, block-wise learning rate, and label smoothing. Our experiments proved that our proposed pipeline boosts the performance significantly, achieving a new state-of-the-art on Vietnamese Hate Speech Detection (HSD) campaign\footnote{https://vlsp.org.vn/vlsp2019/eval/hsd} with 0.7221 F1 score.
\end{abstract}

\IEEEpeerreviewmaketitle
\section{Introduction}





\subsection{Overview}
The rapid growth of the Internet, social media, and community forums have allowed people across the world to connect instantaneously and has revolutionized communication as well as content issues. However, the increase of hate speech on these platforms has drawn significant expenditure from governments, organizations, companies, and researchers. The term ``hate speech'' can be understood as any kind of communication that uses pejorative or discriminatory language with reference to a person or a group based on their religion, gender, ethnicity, nationality, race, colour, descent, or other identity factors. Multiple statistics reports \cite{10.1093/bjc/azz064} and books \cite{shiell2009campus} also show that hate speech and crime are highly correlated and on the rise together.
Since the internet gives people some degree of anonymity, some take this for granted and abuse it to harass others. Calling names, making distasteful comments about one's origin, or simply shaming someone in anyway, are all everyday examples of hate speech that anyone will encounter occasionally. To combat this, a vast number of methods have been studied and developed for automated HSD. This aims to classify textual content into hate and non-hate speech. 

By the end of 2019, social network site users in Vietnam have reached 48 million users. Still, there has been limited available research about Vietnamese HSD; building appropriate countermeasures for hate speech requires detecting and tracing through content. In the case of the Vietnamese language, this task becomes difficult due to the diverse vocabulary and complex grammar. For example, the same subword can have multiple meanings, which is different from Latin word roots. Or, the fact that the same base (sub)word having multiple possible intonations creates many hindrances in the path of language understanding: not only each subword has multiple vastly different meanings, but also this causes an infinite number of combinatorial possibilities of either misspelling or intentional shorthand expressions.


The variety of semantics and grammar in the Vietnamese language has led to a big challenge for automatic hate speech detection. Previous researches of text classification based on traditional machine learning algorithms or training deep learning from scratch is often inefficient. It also requires a lot of effort for pre-processing and assimilating the semantic of words. In addition, recently proposed pre-trained language models have accomplished success in multiple tasks of natural language processing through fine-tuning when integrated with the model of downstream tasks. The emergence of pre-trained language models has contributed to new ideas for solving significant problems. Pre-trained language models are large scale neural network models based on the deep Transformer structure. Their initial parameters are learned through immense self-supervised training, then combined with multiple downstream models to fix special tasks by fine-tuning. Empirical experiment's results show that the downstream tasks’ performances of these kinds of models are usually better than those of conventional models.
\subsection{Our contributions}
In this paper, we investigate many experiments in fine-tuning the pre-trained RoBERTa model for text classification tasks, specifically Vietnamese HSD. We propose a general pipeline and model architectures to adapt the universal language model as RoBERTa for downstream tasks such as text classification. With our technique, we achieve new state-of-the-art results on the Vietnamese Hate Speech campaign, organized by VLSP 2019\footnote{The sixth international workshop on Vietnamese Language and Speech Processing.}.

The main contributions of our paper are as follows:
\begin{itemize}
\item We propose a general pipeline to adopt the universal language model as a pre-trained RoBERTa for the text classification. It includes two steps: (1) Re-training masked language model task on training data of the classification task. (2) Fine-tuning model with a new classification head for the target task.

\item We conduct multiple methods to design model architecture for text categorization task by using the pre-trained RoBERTa model such as PhoBERT\cite{nguyen2020phobert}.

\item A number of training techniques are suggested that can improve the efficiency of the fine-tuning phase in solving data problems. These techniques are practical and can empirically help the model prevent overfitting phenomenon in the absence of training data or data imbalances.

\item From PhoBERT to Vietnamese HSD: We have achieved the new state-of-the-art results on Vietnamese HSD task by utilizing PhoBERT and our proposed method.

\end{itemize}

\subsection{Roadmap}
The rest of the paper is organized as follows. Section 2 provides a brief survey of related work. Next, in Section 3, we introduce our proposed method, the model architecture, and our conducted fine-tuning strategies. Experiments are described in Section 4, including dataset information, data processing method, and some other experimental settings. Section~5 shows our experimental results. Finally, Section 6 presents our conclusions.

\section{Related work}

\subsection{Language Models}
Language modeling has been becoming an essential part of modern NLP field. It helps computers understand human language by digitalizing qualitative information. Early studies on word embeddings try to construct static representations for words, that is, context-independent embeddings. Mikolov\textit{ et al.}~\cite{mikolov2013efficient} introduced novel architectures with an open-source package called \textit{Word2Vec}. The architectures consist of two models: Continuous Bag of Words (CBOW) and Skip-gram model were trained on a huge dataset of 1.6 billion words. GloVe \cite{pennington-etal-2014-glove} is an unsupervised learning algorithm for context-independent word embedding extraction. It first creates co-occurrence matrix of words and then factorizes it to extract dense word vectors. However, GloVe and Word2Vec are both fail in representing rare or out-of-vocabulary words. To mitigate this problem, fastText~\cite{mikolov2018advances} decomposes each word as a sum of character n-grams. This handles unseen words very well because their character n-grams still occur in other words. In contrast to context-independent embeddings, contextualized word embeddings aim to encode word semantics within contexts. Marking a new era of NLP, Bidirectional Encoder Representations from Transformers \cite{devlin2018bert} or BERT achieves new state-of-the-art performance on eleven NLP tasks, outperforms previous best result (with GLUE score of 80.4\%, 7.6\% improvement) and human on SQuAD 1.1 (with 93.2\% accuracy, 2\% higher). The large model consists of 24 Transformer blocks for a total of 340M parameters and was trained on 3.3 billion words corpus. GPT-2 \cite{radford2019language} by OpenAI is even a larger model with 1.5 billion parameters and 48 layers. Being trained on a huge, diverse and well-processed dataset, GPT-2 achieves state-of-the-art results on 7 out of 8 datasets. Facebook research team propose an improved training procedure for BERT, called RoBERTa \cite{liu2019roberta}. The improvements include a ten-times larger dataset, longer training, increased batch size, using byte-level encoding with bigger vocabulary, excluding next sentence predicting task and dynamic masking pattern changing. For Vietnamese language modeling, PhoBERT \cite{nguyen2020phobert} is the current state-of-the-art. The model is based on RoBERTa with two versions ``base'' and ``large''. Both versions outperform previous best on different downstream tasks.

\subsection{Hate Speech Detection (HSD)}
HSD can be understood as a type of text classification. Before the deep learning era, traditional methods had been widely studied for this notorious problem. These approaches require manual feature engineering to encode text sequences in vector form, which then be fed into classifiers such as Random Forests \cite{inproceedings3}, Naive Bayes \cite{6406271, inproceedings2}, and Support Vector Machine \cite{article, inproceedings2, inproceedings3}. \textit{Bag-of-words (BoW)} is a way to represent documents by modeling the occurrence of every word. This has been reported to be a discriminative feature for HSD \cite{article, Burnap2016UsAT, 6406271, hosseinmardi2015detection, 10.1002/asi.21690, inproceedings, waseem-hovy-2016-hateful,  10.5555/2382029.2382139}. However, modeling words suffers from the problem of typing error, which generates noisy words to the corpus. \textit{Character n-gram} divides a word into sub-words in order to suppress typing error parts of the word. In \cite{inproceedings2}, the authors have systematically proved that character-level n-gram representation outperforms 
BoW in abusive language detection. Based on the idea that hate speech usually comes along with negative sentiment, the methods in \cite{10.1145/2362394.2362400, article2, 10.1002/asi.21690} utilize \textit{sentiment analysis} as an evidence for HSD. \textit{Linguistic features} inject additional useful knowledge to the raw text. Xu~\textit{et al.} \cite{10.5555/2382029.2382139} combine Stanford CoreNLP \cite{manning-etal-2014-stanford} POS information with n-gram features. However, the POS tokens did not improve the performance of the classifiers. In contrast, \cite{article, Burnap2016UsAT, 6406271, article2, 10.1145/2872427.2883062} successfully employed dependency relationships in their feature set and report significant performance improvements. Because hate speech usually comes from social media, it is feasible to extract meta-information such as how likely a user will post hateful speech in the future \cite{inproceedings3}, the number of profanity in a user's activity history\cite{dadvar} or user gender \cite{dadvar2012,waseem-hovy-2016-hateful}.

Recent studies are paying great attention to deep learning approaches. It not only boosts performance considerably, but also requires less effort on feature engineering. The input of a deep learning model can simply be an one-hot encoding of text sequences \cite{dllstm, dlcnn, dlcnnlstm, dllstm2}, then meaningful features are learned by Convolutional Neural Networks (CNN) \cite{dlcnn}, Long Short-Term Memory (LSTM) \cite{dllstm, dllstm2} or even a combination of CNN and LSTM \cite{dlcnnlstm}. However, one-hot representation suffers from issues of high dimensionality as its length equals to the vocabulary size. Therefore, a more convenient way is to embed the input into a low-dimensional space. This can be character embeddings \cite{inproceedings2}, comment embeddings \cite{10.1145/2740908.2742760} or text embeddings from a two-phase deep learning model \cite{Yuan2016ATP}.

HSD Shared Task in VLSP Campaign 2019 \cite{vu2020hsd} is a challenge for detecting Vietnamese hate speech on social networks. Our team, the winning team, using logistic regression with ensemble learning n-gram, achieved F1 score of 0.67756 and 0.61971 on public-test and private-test, respectively. The second best team also utilized logistic regression with ensemble learning, however the input includes more feature types: n$-$grams of words, part-of-speech tags, and numeric features. They scored 0.58883 on private-test, that is 3\% lower than the winner. The other teams employed deep learning architectures from CNN, LSTM, RNN to Bi-LSTM, LSTMCNN and Bi-GRU-LSTM-CNN, achieving scores from 0.51705 to 0.58455 on private-test.
\section{Proposed method}
\subsection{RoBERTa-based Network for HSD Task}

\begin{figure}[t]
    \centering
    \includegraphics[width=70mm]{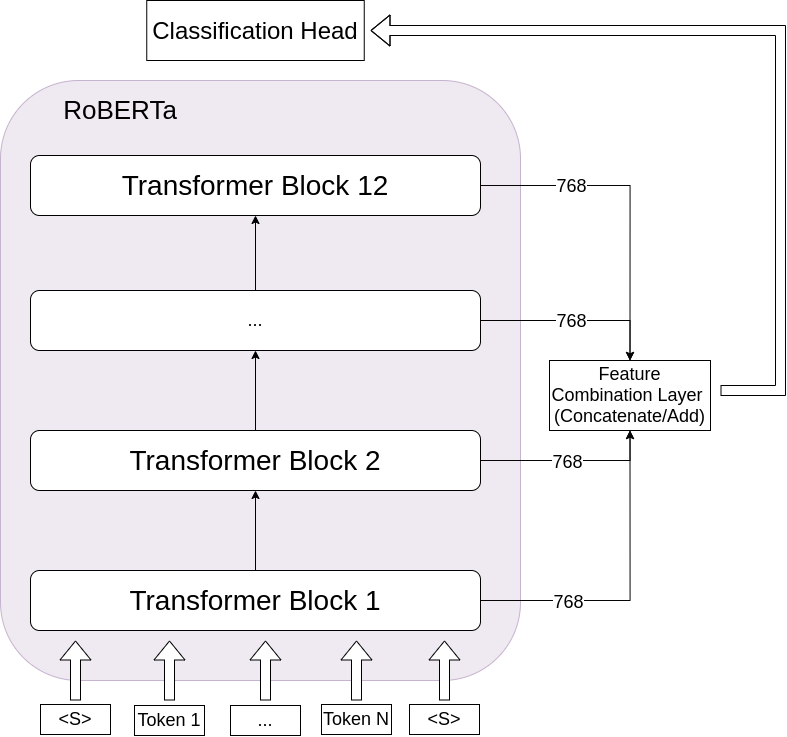}
    \caption{An illustration of our architecture. The input is a tokenized sentence, feeding parallelly to RoBERTa-base. Each of the twelve transformer blocks of the RoBERTa produces a 768-D vector. These vectors are then concatenated to form a long vector for the classification head.  We  investigate on the performance of different combinations of these vectors.}
    \label{fig:cfm}
    
\end{figure}

Our architecture utilizes PhoBERT as a backbone network with some modifications. The output of each Transformer block represents a different semantic level for the inputs. In our experiments, we use different combinations of outputs of those Transformers blocks. The general architecture model is shown in Figure~\ref{fig:cfm}.

The features are combined across outputs of multiple transformer blocks by concatenating or adding are fed to the classification head. A simple classification head is a fully connected network without hidden layers.

\subsection{Proposed Fine-Tuning Strategies}

\begin{figure}
    \centering
    \includegraphics[width=\linewidth]{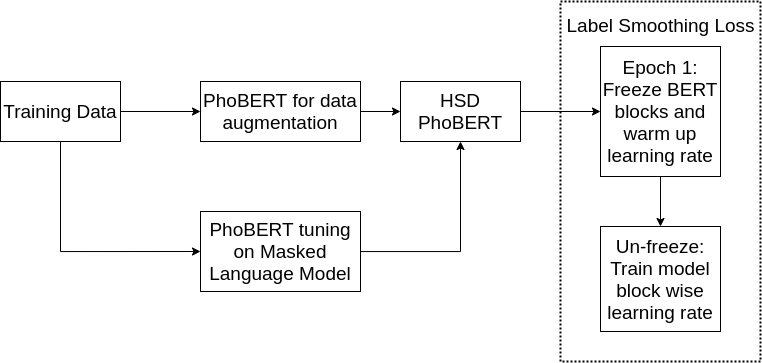}
    \caption{The proposed fine-tuning strategies for Hate Speech detection.}
    \label{fig:pipeline}
\end{figure}

\subsubsection{Fine-Tuning Masked Language Model}
The pre-trained model was fit to a much larger dataset of a completely different domain. Therefore, while it might work very well in general, the vanilla pre-trained model will underperform at our one specific task. This induces the need for us to tune the model to our needs. Therefore, with the existing weights of PhoBERT as a starting point, we train our model with our domain-specific training data on Masked Language Modeling (MLM) task.

\begin{figure}
    \centering
    \includegraphics[width=60mm]{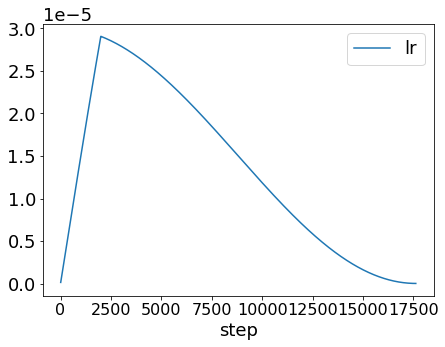}
    \caption{Warm-up learning rate}
    \label{learning_rate}
\end{figure}

Moreover, for such a large model to be trained successfully without either forgetting its good initialization (by gradient descending too far from it) or failing to converge (due to deep models being bad at propagating through further layers), we use a warm-up learning rate scheduling scheme. Originating from paper \cite{howard2018universal} under the name of ``slanted triangular learning rates", the main purpose of this method is to make the model converge more quickly for a suitable region of the parameter space in the beginning process of training.

\subsubsection{Optimization Strategy: Freeze or not Freeze}
Each layer in RoBERTa network captures different levels of context. Specifically, lower layers produce global embeddings for words while the embeddings from upper layers are more context-specific. We would like to preserve the global knowledge while adjusting context representations for our classification model.

For the first few epochs, we only keep the fully connected layers that are responsible for classifying text sequences, and the RoBERTa part is frozen. This allows the model to learn a decent decision for the task. After these epochs, we unfreeze all layers and set different learning rates for different layers: the deeper the layer is, the more learning rate increased. This prevents the model from forgetting the useful global meaning of the words while forcing it to learn the context of the domain.
\subsubsection{Label Smoothing}
For such a large model to be fit on a relatively small set of data, the model tends to become overconfident of its performance, going to the dark side of overfitting. To avoid this, we employ label smoothing,
which softens our one-hot ground truth labels. Specifically, for a model outputting probabilities $y_k$ of $K$ classes, instead of labelling our ground truth with one-hot encoding:
\begin{equation}
y_k=\begin{cases}
1 & \text{ if }k=j\text{ for some }j,\\
0 & \text{ otherwise,}
\end{cases}
\end{equation}
we slightly rebalance the target distribution so that it becomes less ``binarized'' by smoothing the probabilities with, 
\begin{equation}
y'_k = y_k(1-\alpha) + \alpha/K
\end{equation}
for some smoothing parameter $\alpha$. As a result, this technique teaches the model to have some uncertainty in its guesses, and reduces the severity of overfitting. Moreover, since we are fine-tuning on a pretrained model, the original output probability vector of the model is close to an one-hot. This introduces numerical instabilities if the new true label is also one-hot, since with cross-entropy as the loss function, the loss becomes $1\log 0=-\infty$. Thus, being used here, label smoothing acts as a small perturbation in our numerical method, making the training process more stable, helping our model converge better.

\section{Experiments}
\subsection{Dataset}
To perform experiments on hate speech detection task, we use the HSD dataset from the VLSP workshop campaign in 2019 \cite{vu2019hsd}. The dataset includes 25,431 samples; all were crawled from posts and comments on Facebook, and annotated to one of three classes: hate speech (HATE), offensive but not hate speech (OFFENSIVE), neither offensive nor hate speech (CLEAN) by many annotators. The training data consists of 20,345 items with label distribution is showed in Table \ref{table:training_data}.

\begin{table}[t]
\centering
\caption{Training data distribution} 
\def\tablename{table}
\begin{tabular}{| c | c | c | c |}
\hline
 & \textbf{HATE} & \textbf{OFFENSIVE} & \textbf{CLEAN} \\ 
 \hline
Number of sample & 709 & 1,022 & 18,614 \\
\cline{1-4}
\end{tabular}
\label{table:training_data}
\label{tmp}
\end{table}

In general, this is an imbalanced dataset and contains a lot of noisy data. There is an extremely enormous gap between the number of CLEAN data points and those from the other two categories, which lead to bad results for some methods, especially Deep Learning approaches.

Because the dataset was crawled directly from users' posts and comments on social networks, it has some notable properties such as abbreviations, emoji, special characters, foreign language, teen code, typing errors, etc. This has led to a significant challenge because PhoBERT or some other pre-trained language model are often trained on normal clean data such as Wikipedia data or Vietnamese news.

\subsection{Data preprocessing}
We have built a basic preprocessing module to process the raw text before feeding it into the model. It includes Unicode normalization, lowercase conversion, and the replacement of all emoji characters to the \texttt{EMOJI} label. Private personal information also is masked for privacy purposes such as phone numbers and email addresses. Finally, raw text is segmented by a word segmenter. As PhoBERT employed the RDRSegmenter \cite{NguyenNVDJ2018} from VnCoreNLP\footnote{https://github.com/vncorenlp/VnCoreNLP} to pre-process training data of the language model task, it is recommended to use the same word segmenter for PhoBERT-based downstream applications.

\subsection{Experimental Settings}

\subsubsection{Data augmentation}

To partially prevent data imbalance and less data for two HATE and OFFENSIVE labels, we used PhoBERT Masked Language Model as a method to augment data. From the original text, we randomly selected one token, replaced it with \textless mask\textgreater, and used the PhoBERT MLM to fill the mask. Repeated 5 times, we got a brand new sample by fill up 5 tokens \textless mask\textgreater from the original text.

\subsubsection{Model Training}
We fine-tuned PhoBERT with Masked Language Model on HSD dataset with ratio of 15\% masked tokens. As in BERT default settings, batch size is fixed to 32, with 10.000 steps. We chose Adam for the optimization and trained the model on a single GPU with learning rate of \num{3e-5}. Figure~\ref{fig:pipeline} illustrates the overall pipeline. After tuning PhoBERT process completed, we retrain HSD model with one or combination of multiple output features from numerous transformer blocks. 

These features then would be incorporated into a Classification Head, which was designed straightforward with Fully Connected and Dropout layers.  
AdamW \cite{loshchilov2017decoupled} were selected to be the optimizer for BERT blocks and Classification Head with different learning rates. BERT learning rate was \num{1e-5} while classification head learning rate was larger at \num{1e-4}. Except bias and LayerNorm being excluded from weight decay, this ratio was chosen with value 0.01. Num\_warmup\_steps in experiments is equal to \num[fraction-function = \sfrac]{1 / 8} of total steps in 1 epoch.\\

We utilized Stratified K-fold with $k = 10$, the ratio of samples in train dataset and valid dataset were preserved as in the original dataset. Each and every fold was trained for 10 epochs.

\subsubsection{Loss function}
We use Label Smoothing loss which is the combination of Cross-Entropy Loss and Label Smoothing. The smoothing rate is set to 0.2 as this shows prominent after the first few epochs.
We slightly lower loss target values while increasing the target value 0. These so-called soft targets will give lower loss when there is an incorrect prediction and consequently, our model will penalize low entropy predictions or ``confidence penalty'' as mention in \cite{mller2019does} and the section above.

The cross entropy loss function is calculated as follows:
\begin{equation}
\mathcal{L}(\mathbf{x}',\mathbf{x})=-\sum_{i}x'_{i}\log x_i,
\end{equation}
where $\mathbf{x}'$ is the true distribution of any particular data point (one-hot encoded, possibly with label smoothing applied), and $\mathbf{x}$ is the model's predicted distribution. Specifically, for our prediction task with a dictionary of $c$ possible words to choose from, $\underset{0\le i<c}{x_i}$ is the probability that the next word in the sequence is the $i$-th token in the dictionary. The loss for the whole dataset (or batch) is the sum (or mean) of the losses of individual data points.

\subsection{System configuration}
Our experiments are conducted on a computer with Intel Core i7 9700K Turbo 4.9GHz, 32GB of RAM, GPU GeForce GTX 2080Ti, and 1TB SSD hard disk. 
\section{Experimental Results}

\subsection{Evaluation metrics}

Macro F1-score is a common evaluation metrics for classification tasks. F1-score is the harmonic mean of $Precision$ and $Recall$.


\textbf{F1 score:} performance measure for classification
\begin{equation}
F_1=\frac{2}{Recall^{-1} + Precision^{-1}}, 
\end{equation}
where $Precision$ is the number of correct positive results divided by the number of all positive results, and $Recall$ is the number of correct positive results divided by the number of all samples that should have been identified as positive.

\textbf{F1-macro} or macro-averaged F1 score is computed as mean of F1 scores for each class

\begin{equation}
Macro\ F_1=\frac{F_{1\text{HATE}} + F_{1\text{OFFENSIVE}} + F_{1\text{CLEAN}}}{3}
\end{equation}

\subsection{Our results}
Our experiment results are shown in Table \ref{table:experimentresults} and Table \ref{table:experimentresults2}. 

\begin{table}[t]
\centering
\caption{Mean of Macro $F1$ score on Stratified K-fold with k = 10 of difference blocks}
\def\tablename{table}
\begin{tabular}{| c | c |}
\hline
\textbf{Feature blocks} & \textbf{Mean of F1 score} \\ 
 \hline
Layer 6 (only single block) & 0.6854 \\
\cline{1-2}
Layer 12 (only single block)& 0.6978 \\
\cline{1-2}
Layer 3-6 (4 middle blocks)& 0.6855 \\
\cline{1-2}
Layer 9-12 (4 last blocks) & \textbf{0.6989} \\
\cline{1-2}
Layer 1-6 (6 first blocks) & 0.6905 \\
\cline{1-2}
Layer 7-12 (6 last blocks)& \textbf{0.6989} \\
\cline{1-2}
Layer 1-12 (all blocks)& 0.6979 \\
\cline{1-2}
\end{tabular}
\label{table:experimentresults}
\end{table}

Table \ref{table:experimentresults} indicates the mean of Macro $F_1$ score on using feature from different BERT blocks. We take the test on multiple selected blocks including single block (layer 6, 12), middle blocks (layer 3-6), last blocks (layer 6-12, 7-12, 9-12) and all blocks. Our retrieved result with last blocks by far is the top score. Last 4 blocks (layer 9-12) and last 6 blocks (layer 7-12) both achieve the highest F1 score at 0.6989.

Table \ref{table:experimentresults2} demonstrates effectiveness of each and every fine-tuning methods. Each individual technique boosts performance of the model by 0.5-1.5\% in term of F1 score while the combination of these methods significantly enhances the score up to 0.7211, outperforming the winner of this challenge (0.67756 F1 score) and the current best result on the public leaderboard (0.71432 F1 score with a single model). Some samples that our system good and failure classified are shown in Table \ref{table:exampleresults}. It shows that our proposed method is efficient for task of HSD. 

\begin{table}[t]
\centering
\caption{Some samples that our system good and failure classified}
\def\tablename{table}
\begin{tabular}{| c | c | c |}
\hline
\textbf{Text} & \textbf{Model prediction} & \textbf{Truth label} \\ 
 \hline
\foreignlanguage{vietnam}{"học kỳ cuối như đồ thị hình sóng thần"} & CLEAN & CLEAN \\
\foreignlanguage{vietnam}{"lan anh đm ám ảnh vl"} & OFFENSIVE & OFFENSIVE \\
\foreignlanguage{vietnam}{"nguyễn hoàng bách dm a mồm lol à"} & HATE & HATE \\
\cline{1-3}
\foreignlanguage{vietnam}{"hay vat đúng lắm ạ"} & OFFENSIVE & CLEAN \\
\foreignlanguage{vietnam}{"có deo đâu mà xem vl"} & CLEAN & OFFENSIVE \\
\foreignlanguage{vietnam}{"hường lily mặt ngu vl"} & OFFENSIVE & HATE \\
\cline{1-3}
\end{tabular}
\label{table:exampleresults}
\end{table}

\begin{table}[t]
\centering
\caption{Mean of Macro $F1$ score on Stratified K-fold with k = 10 with concatenate of layers 6-12 and our training approach}
\def\tablename{table}
\begin{tabular}{| c | c |}
\hline
\textbf{Proposed training approach} & \textbf{Mean of F1 score} \\ 
 \hline
Cross entropy loss & 0.6922 \\
Label Smoothing loss & 0.7005 \\
\cline{1-2}
Non warm-up learning rate & 0.6989 \\
Warm-up learning rate & 0.7062 \\
\cline{1-2}
Non Fine-tune MLM & 0.6989 \\
Fine-tune MLM & 0.7162 \\
\cline{1-2}
Non Block wise learning rate & 0.7051 \\
Block wise learning rate & 0.7079\\
\cline{1-2}
Combine all the methods & \textbf{0.7211} \\
\cline{1-2}
\end{tabular}
\label{table:experimentresults2}
\end{table}

\section{Conclusion}

In this paper, we have explored and proposed our pipeline to solve the Vietnamese Hate Speech Detection task by using a pre-trained universal language model. By intensively conducting experiments using PhoBERT, we have demonstrated that RoBERTa and our fine-tuning strategy is highly effective in text classification tasks. With our proposed methods, we have achieved new state-of-the-art results on the Vietnamese HSD campaign. For future work, we would like to explore different classification head architectures. Instead of only using the fully connected layer, we will experiment with other architectures for instance LSTM, RNN, and CNN-LSTM on top. 

\section*{Acknowledgment}
This work is partially supported by \textbf{\textit{Sun-Asterisk Inc}}. We would like to thank our colleagues at \textbf{\textit{Sun-Asterisk Inc}} for their advice and expertise. Without their support, this experiment would not have been accomplished. We also express our gratitude to Tiep Vu, founder of AIVIVN\footnote{https://www.aivivn.com/} for supporting us in evaluating results on aivivn.com.

\bibliographystyle{plain}
\bibliography{references}
\end{document}